%% file: main.tex
\documentclass[twocolumn,10pt]{asme2ej}

\usepackage{epsfig} 
\usepackage{amsmath}
\usepackage{amssymb}
\usepackage{bm}
\usepackage[utf8]{inputenc}
\usepackage[english]{babel}
\usepackage{algorithm} 
\usepackage{algpseudocode} 
\usepackage{tabularx}
\usepackage{makecell}

\usepackage{graphicx}
\usepackage{url}
\usepackage{epstopdf}
\DeclareGraphicsExtensions{.eps,.pdf}

\providecommand{\keywords}[1]
{
  \textbf{\textit{Keywords---}} #1
}

\usepackage[super, comma, sort&compress]{natbib} 
\usepackage[normalem]{ulem}

\usepackage{multirow}

\usepackage{hyperref}

\title{A Comparison of Pneumatic Actuators for Soft Growing Vine Robots}

\author{Alexander M. Kübler$^{1,2}$}
\author{Cosima du Pasquier$^{1}$}
\author{Andrew Low$^{1}$} 
\author{Betim Djambazi$^{2}$} 
\author{Nicolas Aymon$^{2}$} 
\author{Julian Förster$^{2}$}
\author{Nathaniel Agharese$^{1}$}
\author{Roland Siegwart$^{2}$}
\author{Allison M. Okamura$^{1}$}

\begin{document}
\maketitle

\let\thefootnote\relax\footnotetext{$^{1}$CHARM Lab, Department of Mechanical Engineering, Stanford University, Stanford, CA 94305, USA. Email: \{cosimad@, andrewlow@alumni., agharese@, aokamura@\}stanford.edu}
\let\thefootnote\relax\footnotetext{$^{2}$Autonomous Systems Lab, Department of Mechanical and Process Engineering, ETH Zürich, 8001 Zürich, Switzerland. Email: \{akuebler@, betimd@, naymon@, fjulian@, rsiegwart@\}ethz.ch}

\input{chapter_0_abstract}
\input{chapter_1_introduction}
\input{chapter_2_background}
\input{chapter_3_method}
\input{chapter_4_model}
\input{chapter_5_results}

\input{chapter_6_demonstrator}
\input{chapter_7_conclusion}

\section*{Conflict of interest}
The authors state that there is no conflict of interest.

\begin{acknowledgment}
The authors thank Zhenisbek Zhakypov for help establishing the testing setup, and Pascal Auf der Maur and Patricia Hörmann for help testing the demonstrator.
The project was supported in part by the U.S. Department of Energy, National Nuclear Security Administration, Office of Defense Nuclear Nonproliferation Research and Development (DNN R\&D) under subcontract from Lawrence Berkeley National Laboratory; the United States Federal Bureau of Investigation contract 15F06721C0002306; National Science Foundation grant 2024247; and the DDPS, armasuisse S+T, Swiss Drones and Robotics Centre (SDRC).
\end{acknowledgment}

\bibliographystyle{unsrtnat}
\bibliography{references}

\end{document}

%% file: chapter_0_abstract.tex
\section*{Abstract}

Soft pneumatic actuators are used to steer soft growing ``vine'' robots while being flexible enough to undergo the tip eversion required for growth. In this study, we compared the performance of three types of pneumatic actuators in terms of their ability to perform eversion, quasi-static bending, dynamic motion, and force output: the pouch motor, the cylindrical pneumatic artificial muscle (cPAM), and the fabric pneumatic artificial muscle (fPAM).
The pouch motor is advantageous for prototyping due to its simple manufacturing process. The cPAM exhibits superior bending behavior and produces the highest forces, while the fPAM actuates fastest and everts at the lowest pressure. We evaluated a range of dimensions for each actuator type. Larger actuators can produce more significant deformations and forces, but smaller actuators inflate faster and can evert at a lower pressure. Because vine robots are lightweight, the effect of gravity on the functionality of different actuators is minimal. We developed a new analytical model that predicts the pressure-to-bending behavior of vine robot actuators.
Using the actuator results, we designed and demonstrated a 4.8\,m long vine robot equipped with highly maneuverable 60x60\,mm cPAMs in a three-dimensional obstacle course. The vine robot was able to move around sharp turns, travel through a passage smaller than its diameter, and lift itself against gravity.

\keywords{Soft growing robots, pneumatic actuators, modeling}
\newpage

%% file: chapter_1_introduction.tex
\section{Introduction}
\label{Introduction}

\begin{figure}[t]
    \centering
    \includegraphics[width=0.5\textwidth]{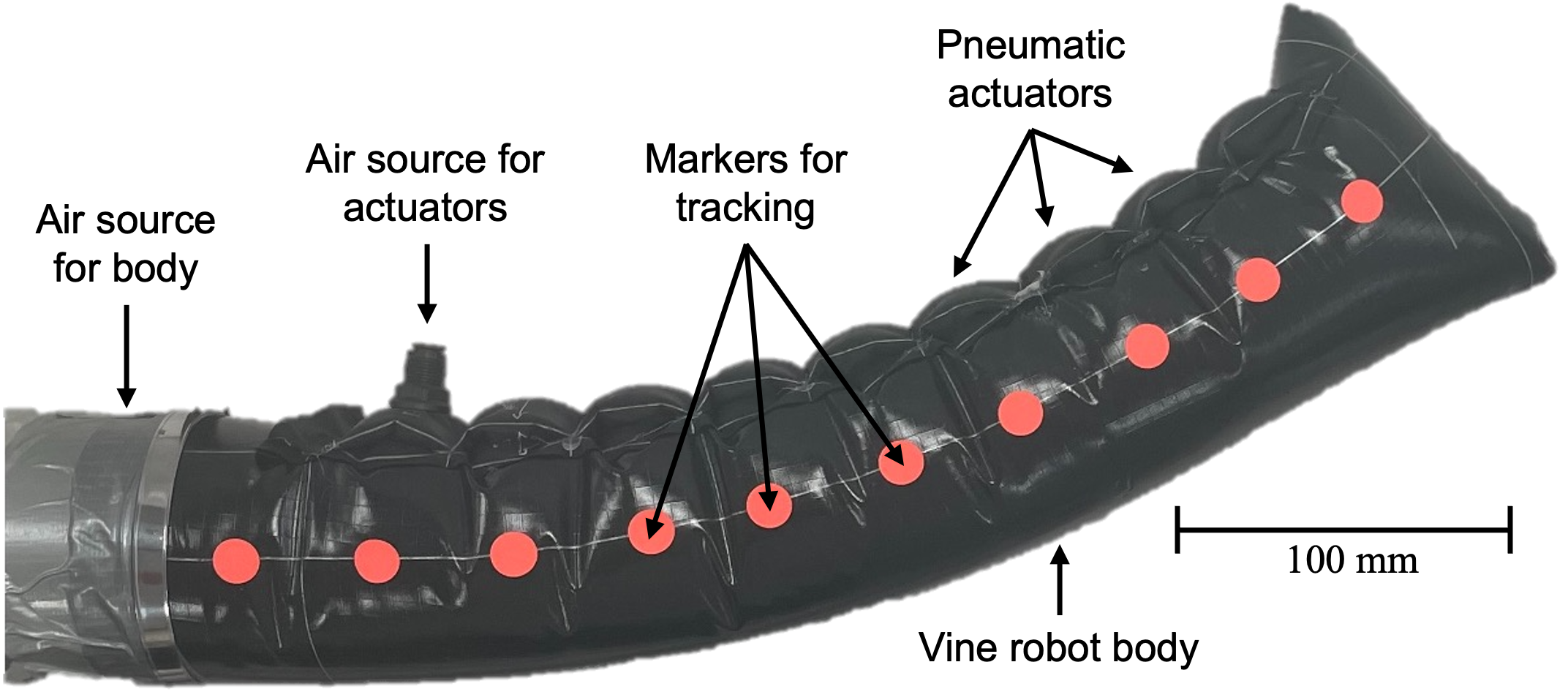}
    \caption{
    This work analyzes pneumatic actuators for vine robots: pouch motors, cylindrical pneumatic artificial muscles (cPAM), and fabric pneumatic artificial muscles (fPAM). The depicted vine robot uses cPAMs of width 60\,mm and length 40\,mm. Its body acts as an inflated beam: when the actuators contract, the vine robot bends.
    }
    \label{fig:first_page}
\end{figure}

Tip-everting soft robots can reach and explore environments inaccessible by classic, rigid robots. When faced with a narrow or cluttered environment, such as a pipe or debris, rigid robots lack the flexibility to cope with unfamiliar and unpredictable situations. Soft growing robots, which extend through an eversion process and thus do not need to slide relative to their surroundings, have been developed to navigate these unmapped environments \cite{Hawkes2017AGrowth,Coad2020VineExploration}.

A soft growing robot, referred to as a ``vine robot'', consists of a cylindrical sleeve of thin, flexible material inverted into itself. It elongates at the tip through eversion when pressurized pneumatically or hydraulically \cite{Luong2019EversionReefs,Blumenschein2020DesignRobots}. This locomotion principle is frictionless with respect to the environment because the vine robot's body does not move relative to its environment.
Variations of vine robots have been tested in archaeological sites \citep{Coad2020VineExploration}, for search and rescue purposes \citep{DerMaur2021RoBoa:Applications}, underwater with the potential of handling coral reefs \citep{Luong2019EversionReefs}, for haptic feedback \citep{Agharese2018HapWRAP:Device}, and in medical applications \citep{Hwee2021AnDevice,Berthet-Rayne2021Mammobot:Detection}.
\begin{figure*}[!ht]
    \centering
    \includegraphics[width=\linewidth]{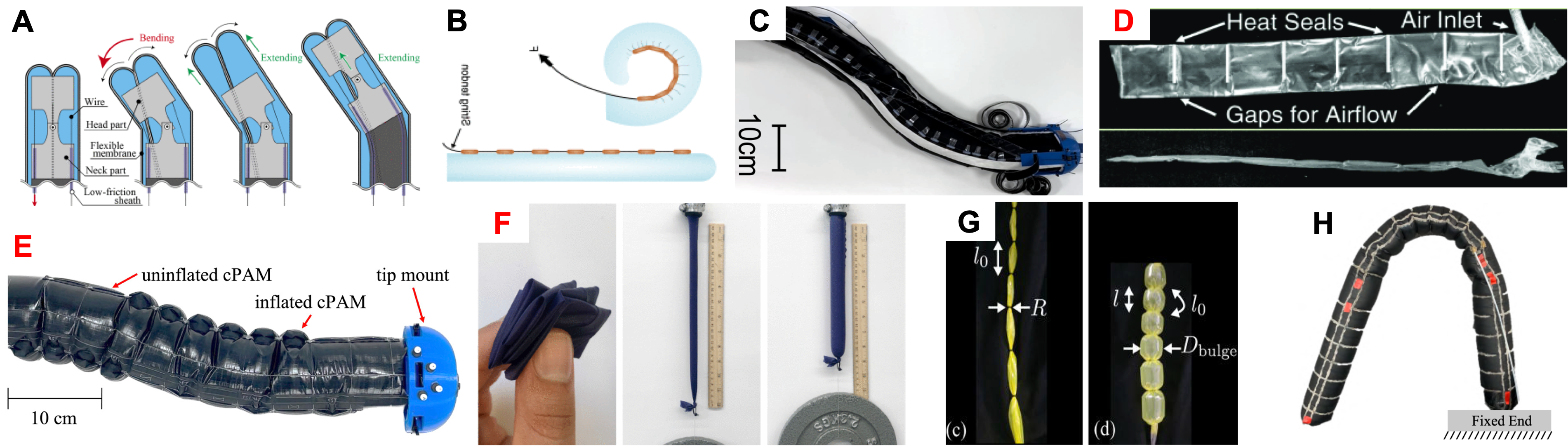}
    \caption{Steering mechanisms for tip everting soft growing robots: A. Rigid internal steering device from \citet{Takahashi2021EversionFunction}. B. Tendon steering from \citet{Gan20203DRobots}. C. Tendon steering with shape-locking using velcro from \citet{Jitosho2023PassiveRobots}. D. Pouch motor design from \citet{Coad2020VineExploration}. E. cPAM design from \citet{Kubler2022ASteering}, similar to the foldPAM introduced by \citet{wang2023folded}. F. fPAM design from \citet{Naclerio2020SimpleMuscle}. G. sPAM design from \citet{Greer2019AExtension}. H. Integrated pouch design from \citet{Abrar2021HighlyStructure}. The steering mechanisms D, E, and F, highlighted in red, are characterized in this work.
    }
    \label{fig:background_act_types}
\end{figure*}
In most applications, controlling the vine robot's path is essential for navigation or to avoid contacting its environment. As shown in the examples in Figure \ref{fig:first_page} and \ref{fig:background_act_types}, steering has been achieved through rigid internal steering devices \cite{Takahashi2021EversionFunction,Haggerty2021HybridCapabilities,DerMaur2021RoBoa:Applications}, using tendons \cite{Gan20203DRobots,Blumenschein2022GeometricRobots}, and with soft actuators such as pouch motors \cite{Niiyama2015PouchDesign,Coad2020VineExploration} and fabric pneumatic artificial muscles (fPAMs) \cite{Naclerio2020SimpleMuscle}.\\
There are limited experimental comparisons and models to guide vine robot designs in actuator selection and optimization. Our objective in this work is to provide thorough experimental characterization and modeling of previous and new vine robot steering actuators.\\
The contributions of this work are the following: First, we establish a benchmark testing procedure and use it to compare three pneumatic actuators: pouch motors, fPAMs, and cylindrical pneumatic artificial muscles (cPAMs), a new type of pneumatic actuator \cite{Kubler2022ASteering, wang2023folded}. The performance of the actuators is compared using four criteria: eversion pressure, quasi-static bending, dynamic reaction time, and lateral force. Second, through exhaustive testing of a range of dimensions of these vine robot actuators, we establish a design heuristic to select the appropriate actuator type and its dimensions based on the use case. Third, we introduce pressure-to-bending models for all three types of actuators. Fourth, based on the results, we design an improved 4.8\,m long vine robot and demonstrate its capabilities in an obstacle course.

%% file: chapter_2_background.tex
\section{Background} \label{Background}

\subsection{Internal Steering Devices}
Internal steering has been shown with rigid devices (Figure \ref{fig:background_act_types}.A) at the vine robot's tip using wires \cite{Takahashi2021EversionFunction,Haggerty2021HybridCapabilities} or 3D printed pneumatic actuators \cite{DerMaur2021RoBoa:Applications}.
They can steer precisely and maneuver heavy payloads but are rigid and heavy, contradicting some key advantages of vine robots, such as the lightweight structure and the possibility to move through holes smaller than the vine robot's diameter.

\subsection{Tendon Steering}
Steering with tendons (Figure \ref{fig:background_act_types}.B) is achieved by routing cables through rigid stoppers attached to the surface of a vine robot \cite{Gan20203DRobots,Blumenschein2022GeometricRobots}. As the robot grows, its tip can be steered by pulling on one or several tendons. The rigid stoppers, usually short sections of a hard plastic tube, provide a controlled limit to the deformation. Tendons thus provide repeatable, precise, and reversible shape changes. However, the friction between tendons and stoppers increases with robot length and limits the deformation range \cite{Gan20203DRobots}.

\subsection{Pneumatic Actuators}
Pneumatic actuation is usually achieved using one of three actuator types: pouch motors, cylindrical pneumatic artificial muscles (cPAMs), or fabric pneumatic artificial muscles (fPAMs). They are attached to the exterior of a vine robot and contract upon pressurization. 2D steering is achieved by attaching two antagonistic sets of actuators to the robot and 3D steering is achieved by attaching three or more actuator sets.
Pouch motors (Figure \ref{fig:background_act_types}.D) are short rectangular discrete actuators. They are manufactured by gluing, welding, or heat sealing rectangles from two layers of inextensible fabric \cite{Niiyama2015PouchDesign,Coad2020VineExploration}. When inflated, they contract into a cylinder-like shape.
cPAMs (Figure \ref{fig:background_act_types}.E) are cut of the same inextensible fabric as pouch motors, but their additional fold on the sides allow them to achieve a near-perfect cylinder and higher contraction ratio than pouch motors \cite{Kubler2022ASteering}. They are also welded directly onto the vine robot's body, facilitating eversion and reducing actuator delamination risk. The foldPAM \cite{wang2023folded} design uses the same folding mechanism as the cPAM. It is designed as a stand-alone actuator where the amount of fold can actively be changed.
The fPAM (Figure \ref{fig:background_act_types}.F) consists of a single cylindrical tube made of anisotropic fabric. Its axis runs parallel to the vine's growth axis \cite{Naclerio2020SimpleMuscle}. fPAMs use the principle of McKibben muscles with an extensible fabric. 
Another pneumatic actuator is the sPAM \cite{Greer2019AExtension} (Figure \ref{fig:background_act_types}.G). Like the fPAM, it consists of a tube but is divided into several pouches that inflate to a bulge. In contrast to the fPAM, it uses an inextensible material. Because of their difficult attachment and large volume due to the balloon-like shape, they are less suited for everting vine robots. \citet{Abrar2021HighlyStructure} introduced an actuator that is directly integrated into the vine robot like the cPAM but consists just of a rectangular pouch like the pouch motor (Figure \ref{fig:background_act_types}.H).

\subsection{Shape-Locking}
Tendon and pneumatic steering can be combined with shape-locking mechanisms to preserve shape change. \citet{Wang2020AShape-locking, Jitosho2023PassiveRobots} use tendon actuation to initiate bending. \citet{Wang2020AShape-locking} guide two to three smaller diameter tip-extending bodies from the base of their main vine robot to modulate stiffness and lock the deformation of a variable body length.
\citet{Jitosho2023PassiveRobots} use velcro straps controlled through a tip mount to fix multi-bend shapes (Figure \ref{fig:background_act_types}.C).
\citet{Kubler2022ASteering} use a tip mount that houses permanent magnets. It selectively opens valves that connect cPAMs to a pressure supply line as the valves move through the tip mount during eversion. After moving through the tip mount, the valves close again and lock the pressure in the corresponding cPAM.

\begin{figure*}[t]
    \centering
    \includegraphics[width=\linewidth]{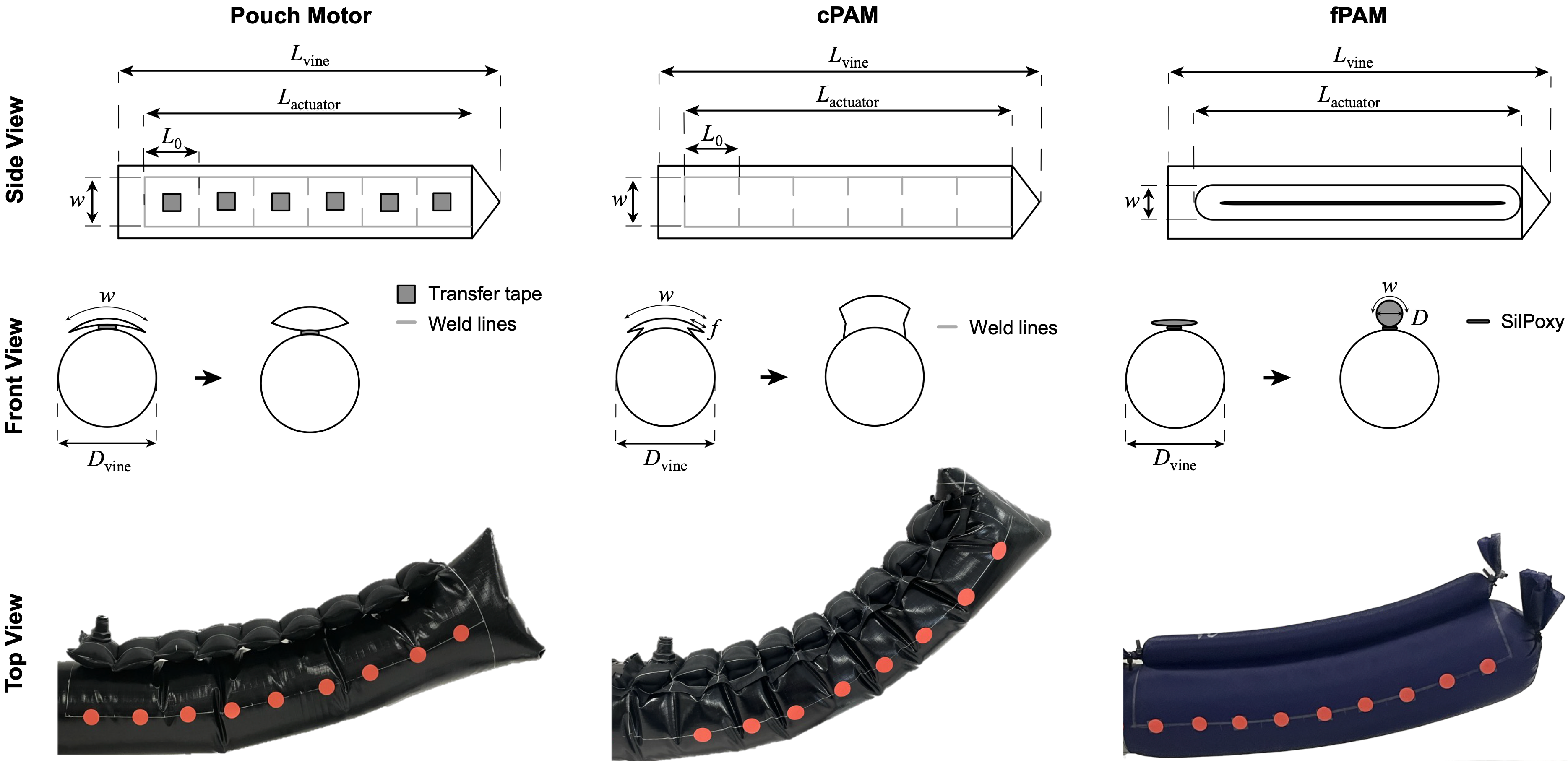}
    \caption{Vine robot fabrication and dimensions overview. Top to bottom: Schematic side and front view with main dimensions, top view of the actuated vine robot. A. Line of welded pouch motors, taped onto the vine robot body. B. Line of welded cPAMs, directly welded onto the vine robot body. C. fPAM, glued using SilPoxy.}
    \label{fig:drawing}
\end{figure*}

\subsection{Modeling of Pneumatic Actuators for Vine Robots}

When using internal steering devices and tendons to steer a vine robot, the orientation of the tip can be calculated geometrically. For pneumatic actuators, the relationship between pressure and deformation is less straightforward because it relies on the contraction through pressure and on complex geometric changes that vary with actuator type and dimension. 
It can either be simulated using finite element modeling as shown by \citet{pasquier2023fem}, or modeled analytically. \\
The analytical model for pouch motors is based on the conservation of energy and the principle of virtual work. Work is defined as a contraction function, determined by an increase in volume \cite{Niiyama2015PouchDesign}. The model assumes a completely inextensible material and an ideal pouch motor without constrained sides. It shows a good force-strain correlation for the linear contraction of isolated actuators. This model approach can also be used for cPAMs because they behave similarly to an ideal pouch motor.
\citet{Naclerio2020SimpleMuscle} presented an analytical model for the fPAM.
Using material parameters, they relate an increase in pressure to the contraction force, based on previous models on the McKibben muscle.
Their model performs well in a comparison of the force-strain relationship with test data.
Both models only represent the linear contraction of the actuator itself, not the bending of the vine robot.
\citet{Abrar2021HighlyStructure} use the ideal pouch motor model for a vine robot with pouches integrated within the vine robot body. They assume static equilibrium force conditions to relate the actuator pressure with the position of the vine robot tip. The model differs
from the experimental data. \\
In this paper, we expand on the linear contraction models from \citet{Niiyama2015PouchDesign, Naclerio2020SimpleMuscle} and the static equilibrium force assumption from \citet{Abrar2021HighlyStructure}. We adapt the models to account for the geometric conditions and varying dimensions of the pouch motor, cPAM, and fPAM.

%% file: chapter_3_method.tex
\section{Fabrication Methods}
\label{Method}

We fabricated vine robots with uniform dimensions for all three actuator types. To ensure a fair comparison, each vine robot has a diameter of ${D_\text{vine}\,=\,80\,\text{mm}}$ because the bending performance depends on the diameter of the vine robot. The length of all vine robots was fixed at ${L_\text{vine}\,=\,420\,\text{mm}}$ and the actuator line length at ${L_\text{actuated}\,=\,360\,\text{mm}}$. We manufactured a vine robot with one actuator line for every actuator type and dimension. This allows for two-dimensional bending in one direction. A second line can achieve two-dimensional bending in two directions, and a third line can achieve three-dimensional steering. For the case of multiple actuator lines, each one can be investigated independently, and the combined bending can be calculated as a superposition. Therefore, the following actuator investigation also holds for the case of multiple actuator lines. 
Figure \ref{fig:drawing} shows schematic drawings and Table \ref{tab:dimensions} states the fabricated dimensions.

\subsection{Pouch Motor} 
The vine robot with pouch motors comprises a sequence of rectangular pouch motors. Both the vine robot body and the pouch motors are made of 70 Denier ripstop nylon with a one-sided TPU (thermoplastic polyurethane) coating (Quest Outfitters, Sarasota, Florida, United States). A line of pouch motors is formed by welding two layers of material together in a rectangular shape using a Vetron 5064 ultrasonic welder (Vetron Typical GmbH, Kaiserslautern, Germany).
The pouch motors are connected to the vine robot body using adhesive transfer tape (3M, Saint Paul, Minnesota, United States), located at the center of the pouch motor.

\begin{table*}[ht]
\caption{Tested dimensions and materials of vine robots with pouch motors, cPAMs, and fPAMs. Crossed-out numbers indicate geometrically infeasible dimensions and dimensions that do not create any bending.}
\label{tab:dimensions}
\small
\begin{center}
\renewcommand{\arraystretch}{1.0}
\begin{tabular}{llccccccccccccc}
\hline
\multicolumn{1}{c}{Dimensions (mm)} & \vline & \multicolumn{2}{c}{} & \multicolumn{3}{c}{$L_{\text{vine}} = 420$} &  \multicolumn{3}{c}{$L_{\text{actuator}} = 360$} & \multicolumn{3}{c}{$D_{\text{vine}} = 80$}  \\
\hline
\hline
\multicolumn{1}{c}{Actuator Type} & \vline & Material & \vline & \multicolumn{10}{c}{Actuator Dimensions (mm)} \\
\hline
\multicolumn{1}{c}{\multirow{2}{*}{Pouch Motor}}   & \vline & 70 Denier ripstop nylon            & \vline & $w$ & \vline & 20      &\sout{20}&\sout{20}& 40      & 40     &    40  & 60      & 60     & 60     \\
                                  & \vline & (one-sided TPU coating)            & \vline & $L_0$ & \vline & 20      &\sout{40}&\sout{60}& 20      & 40     & 60     & 20      & 40     & 60     \\
                                
\hline
\multicolumn{1}{c}{\multirow{3}{*}{cPAM}}          & \vline & \multirow{3}{*}{\shortstack{70 Denier ripstop nylon \\ (one-sided TPU coating)}}            & \vline & $w$ & \vline & 20      &\sout{20}&\sout{20}& 40      & 40     &\sout{40}& 60      & 60     & 60     \\
                                  & \vline &            & \vline & $L_0$ & \vline & 20      &\sout{40}&\sout{60}& 20      & 40     &\sout{60}& 20      & 40     & 60     \\
                                  & \vline &                                    & \vline & $f$ & \vline & 8       &        &        & 8       & 16     &      & 8       & 16     & 24     \\
\hline
\multicolumn{1}{c}{\multirow{2}{*}{fPAM}}          & \vline & 30 Denier ripstop nylon                      & \vline & $w$ & \vline & \multicolumn{3}{c}{20}    & \multicolumn{3}{c}{40}    & \multicolumn{3}{c}{60}    \\
                                  & \vline & (stretchable, two-sided silicon coating) & \vline & $D$ & \vline & \multicolumn{3}{c}{12.73} & \multicolumn{3}{c}{25.46} & \multicolumn{3}{c}{38.20} \\
\hline
\end{tabular}
\end{center}
\end{table*}

\subsection{cPAM}
The line of cPAMs, made from the same material as the pouch motor, is integrated directly into the vine robot body, where the vine robot body functions as the lower layer of the cPAM. Material is folded on either side to form two additional layers, enabling the cPAM to adopt the shape of a cylinder upon inflation. This feature distinguishes the cPAM from a pouch motor, as the latter is restricted on its sides, which hinders bending. The fold length $f$ is related to the diameter $D$ of the cylindrical shape in the inflated state, which is associated with the length $L_0$ of the cPAM \cite{Kubler2022ASteering}.
\begin{gather}
    f = \frac{1}{2} D = \frac{1}{\pi} L_0
\end{gather}
Moreover, the fold length cannot exceed half the width of the cPAM:
\begin{gather}
    f \le \frac{1}{2} w
\end{gather}
This constraint renders dimensions with a greater length than width impractical. To simplify the manufacturing process, we adopted a standard fold length $f$ of 8\,mm per 20\,mm cPAM length $L_0$. The cPAM is fabricated through ultrasonic welding over the two and four layers.

\subsection{fPAM}
The fPAM is constructed from thin, stretchable 30 Denier ripstop nylon with a two-sided silicon coating (Rockywoods Fabrics, Loveland, Colorado, United States). The actuator comprises a single cylindrical tube affixed to the vine robot body, made of the same silicon-impregnated ripstop nylon, using SilPoxy adhesive (Reynolds Advanced Materials, Broadview, Illinois, United States). The fabrication process follows a similar procedure as described by \citet{Naclerio2020SimpleMuscle}. Pre-stretching the fPAM before gluing it onto the vine robot body results in stronger bends by enabling higher contractions of the actuator. Unlike the pouch motor and cPAM, only the diameter $D$ of the fPAM can be modified. It is directly correlated to the width $w$ in the uninflated state:
\begin{gather}
    D = \frac{2}{\pi} w
\end{gather}

\section{Experimental Methods}
\begin{figure}[t]
    \centering
    \includegraphics[width=0.5\textwidth]{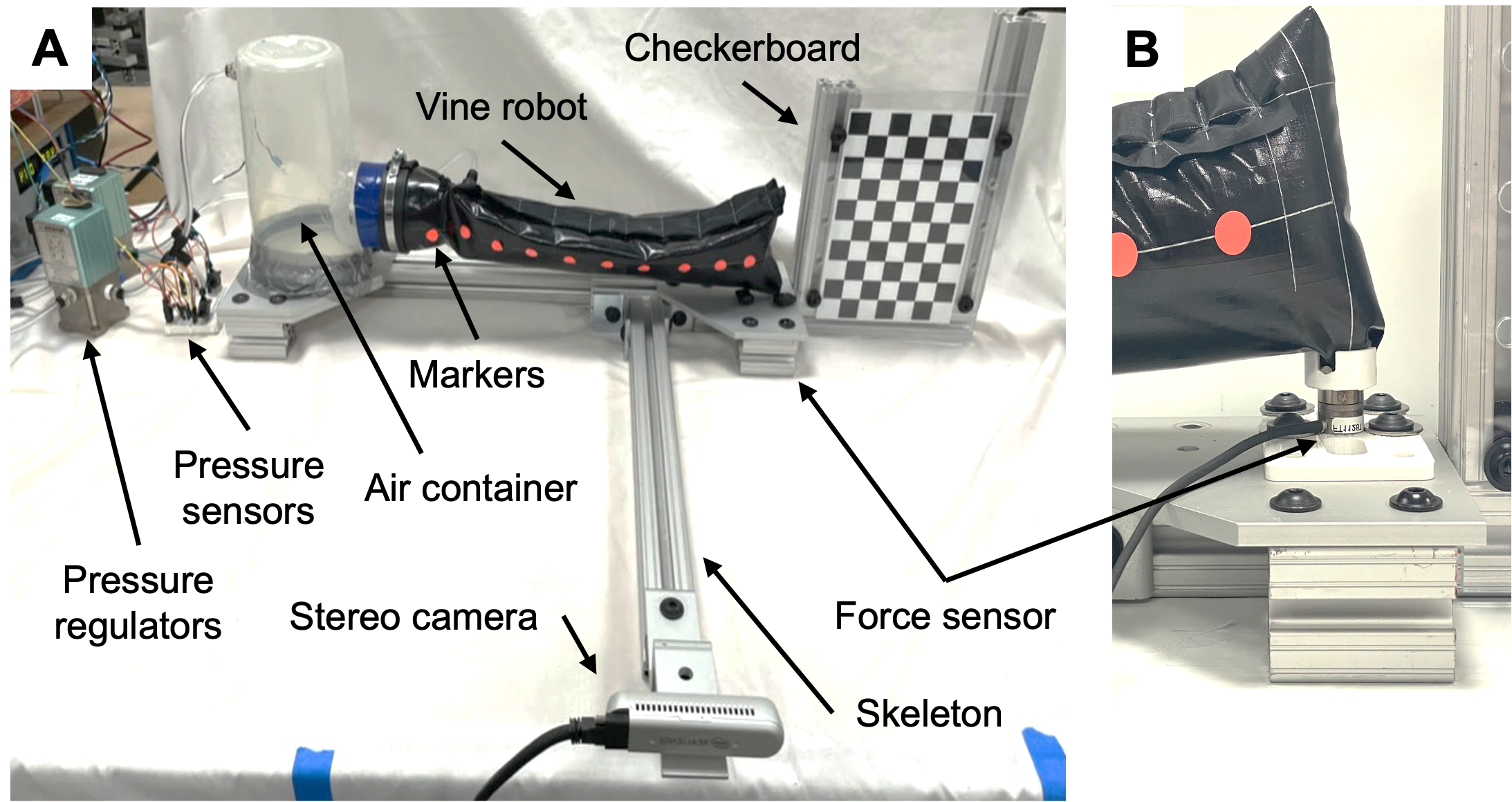}
    \caption{A. Measurement setup for bending, dynamic motion, and force tests. A stereo camera, calibrated by the checkerboard, tracks the deformation of the vine robot using the orange markers. The vine robot body is connected to the air container. Two pressure regulators control the pressure in the vine robot body and the actuators, two pressure sensors sense the actual pressure. Everything is connected by a skeleton made of aluminum profiles to allow for horizontal and vertical orientation. B. The force sensor in detail view.}
    \label{fig:setup}
\end{figure}

We conducted multiple experiments to assess the performance of the three types of actuators with various dimensions. Each test was performed five times to ensure repeatability.
The experimental setup depicted in Figure \ref{fig:setup} is designed to evaluate a fully everted vine robot with a single line of actuators. The setup employs a Realsense D415 camera (Intel, Santa Clara, California, United States) that captures RGB and depth images and is calibrated using a checkerboard. Markers are placed on the vine robot to track its shape and movement. Two QB3 pressure regulators control the pressure: QB3TANKKZP6PSG for the actuator pressure and QB3TANKKZP10PSG for the vine robot pressure (Proportion-Air, McCordsville, Indiana, United States). Two MPX4250AP pressure sensors (NXP Semiconductors, Eindhoven, Netherlands) measure the actual pressure in the vine robot body and the actuators. A Nano-17 6-degree-of-freedom force-torque sensor (ATI Industrial Automation, Apex, North Carolina, United States) positioned at the tip of the vine robot measures the force output. All setup components are connected using a skeleton made of aluminum profiles, enabling horizontal and vertical orientations for testing with or without gravity.

\subsection{Eversion}
This test investigates the pressure required for continuous eversion of the vine robot. The vine robot is turned inside itself with the actuator attached. We apply a pressure to the vine robot body, using the air container and pressure regulator, and inspect if the vine robot everts continuously. If the pressure is not sufficient to achieve eversion, we increase the pressure inside the vine robot body in steps of 0.1\,kPa. Continuous eversion is achieved when the vine robot everts over its full length without stopping. We performed this test for uninflated actuators and actuators inflated to 15\,kPa.

\subsection{Quasi-Static Bending}
\label{method_bending}
This experiment establishes a relationship between the pressure in the actuator and the bending of the vine robot. We maintained a constant pressure of 1.75\,kPa inside the vine robot body. At this pressure, all examined vine robots can evert with uninflated actuators. We slowly increased the pressure (0.65\,kPa/s) in the actuator from 0\,kPa to 40\,kPa to eliminate dynamic effects. For some larger actuators the maximum pressure was reduced to avoid damage (see first row in Figure \ref{fig:results} for the pressure range of each actuator). We measured the bending per length, which is calculated by dividing the angular deflection between two segments (pairs of markers) by the initial length of one segment. The segment length is $2L_0$ for actuators with $L_0 = 20\,\text{mm}$ and $L_0$ for longer actuators. The final bending per length value was obtained by averaging the values across all segments.
The tests were performed in a horizontal and vertical configuration to investigate the influence of gravity.

\subsection{Dynamic Motion}
\label{method_dynamic}
This test uses the same experimental configuration as the quasi-static bending test. But instead of slowly increasing the pressure, we applied a step input to the pressure regulator, which aims to pressurize the actuator as quickly as possible. We inflated the actuators from 0\,kPa to the target pressure value, waited until the vine robot reached a steady state, and then commanded the pressure back to 0\,kPa. We repeated this process for ten different target pressure values for each actuator, equally distributed in the pressure range of each actuator. The pressure range is the same as in Section \ref{method_bending}. We evaluated the dynamic response of the actuators by measuring the 10-90\% rise and fall times of the bending per length.
Again, the tests were performed horizontally and vertically.

\subsection{Force Output}
We measured the force output of actuated vine robots. We fixed the front of the vine robot to the 6-degree-of-freedom force-torque sensor and set the pressure in the vine robot body to 1.75\,kPa. We measured the force output for ten different pressure values for each actuator (same pressure values as in Section \ref{method_dynamic}). The lateral force (perpendicular to the vine robot's growing axis) is our primary metric because its direction is of particular interest for manipulation and steering purposes.

\begin{figure*}[t]
    \centering
    \includegraphics[width=\linewidth]{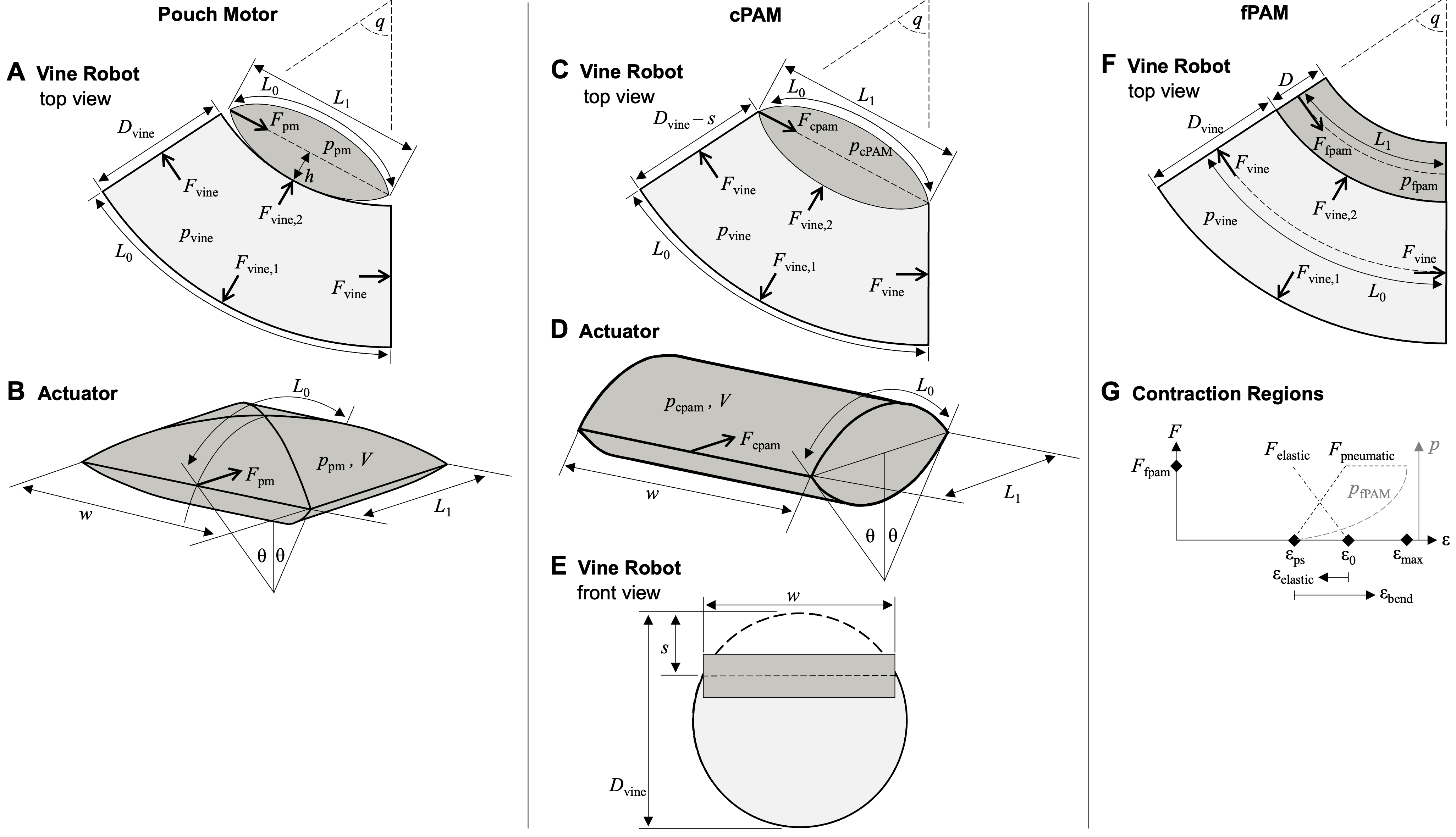}
    \caption{Modeling of the vine robot's pressure-to-bending relationship for different actuator types; schematics indicate the dimensions, applied pressures, and acting forces. A.-B. Pouch Motor: A. Top view of a vine robot segment with an attached pouch motor. B. Single pouch motor modeled as an ideal pouch motor with corrected volume. C.-E. cPAM: C. Top view of a vine robot segment with an integrated cPAM. D. Single cPAM modeled as an ideal pouch motor. E. Front view of a vine robot with an integrated cPAM and resulting diameter correction. F.-G. fPAM: F. Top view of a vine robot segment with an attached fPAM, G. Different contraction regions of the fPAM actuator with corresponding pressure (grey) and acting forces (black) from an elastic, pre-stretched state to a fully inflated state with maximal bending. $\epsilon$ defines the general contraction of the fPAM,
    $\epsilon_\text{elastic}$ defines the elastic contraction, and $\epsilon_\text{bend}$ defines the contraction related to the bending of the vine robot. $\epsilon = 0$ indicates the theoretically maximum pre-strech, $\epsilon = \epsilon_\text{ps}$ the pre-strech when integrated into the vine robot and not inflated, $\epsilon = \epsilon_0$ the contraction without pre-stretch, and $\epsilon = \epsilon_\text{max}$ the maximum contraction. The pressure curve qualitatively shows the relationship between pressure and contraction.}
    \label{fig:modeling}
\end{figure*}

%% file: chapter_4_model.tex
\section{Modeling Methods}
We developed static analytical pressure-to-bending models to predict the bending of each actuator type at specific pressures. We expanded existing linear contraction pouch motor \cite{Niiyama2015PouchDesign} and fPAM \cite{Naclerio2020SimpleMuscle} models. Our models neglect gravity because of the lightweight vine robot design.
Figure \ref{fig:modeling} shows the geometries and forces acting on each  type of actuator.
The core idea is similar for all actuator types, namely that the axial force stretching the vine robot body is in equilibrium with the force produced by the actuator. The axial force of the vine robot can be calculated using the pressure in the vine robot $p_\text{vine}$ and its diameter $D_\text{vine}$:
\begin{equation}
    F_\text{vine} = \frac{\pi}{4} D_\text{vine}^2 p_\text{vine}
\end{equation}
Including the forces on the side of the vine robot has a small impact but complicates the model.
Therefore, we make the simplifying assumption that the forces on the side of the vine robot are equal and cancel each other out, even if the bending of the vine robot is large:
\begin{equation}
    F_\text{vine,1} = F_\text{vine,2}
\end{equation}
The equations vary for the different actuator types because of the diverse geometric conditions and the different linear contraction models for different types of actuators.

\subsection{Pouch Motor}
This model builds on the linear contraction model for an ideal pouch motor proposed by \citet{Niiyama2015PouchDesign}. Because the constrained sides reduce the cross-sectional area of an actual pouch motor, we introduce the volume multiplier $\alpha = \frac{2}{3}$ to account for the decreased volume. The value was derived by examining different geometries using CAD (computer-aided design). We then calculate the pouch motor force $F_\text{pm}$ following the methodology proposed by \citet{Niiyama2015PouchDesign}, which considers the uninflated length $L_0$, width $w$, pressure $p_\text{pm}$, and deformation state defined as the central angle $\theta$ of the circular segment of a pouch motor:
\begin{gather}
    F_\text{pm} = \alpha L_0 w p_\text{pm} \frac{\cos(\theta)}{\theta}
    \label{pm1}
\end{gather}
The same model obtains the relationship between the uninflated length $L_0$ and the contracted length $L_1$:
\begin{gather}
    L_1 = L_0 \frac{\sin(\theta)}{\theta}
    \label{pm2}
\end{gather}
The axis of contraction passes through the center of the pouch motor. Since the pouch motor is mounted on top of the vine robot, as inflation increases, this axis moves away from the vine robot by the distance $h$:
\begin{gather}
    h = \frac{L_1}{2} \tan\left(\frac{\theta}{8}\right)
    \label{pm3}
\end{gather}
Considering the bending $q$ of the vine robot relates the axial vine robot force with the pouch motor force:
\begin{gather}
    F_\text{vine} \sin(q) = F_\text{pm} \sin\left(\frac{q}{2}\right)
    \label{pm4}
\end{gather}
We establish a geometric relationship between the uninflated and contracted lengths and the bending $q$ of the vine robot:
\begin{gather}
    L_0 = \frac{q L_1} {2 \tan(\frac{q}{2})} + q (D_\text{vine}+h)
    \label{pm5}
\end{gather}
Inserting Equations \ref{pm1}, \ref{pm2}, and \ref{pm3} into Equations \ref{pm4} and \ref{pm5}, we input the pouch motor pressure $p_\text{pm}$ and solve Equations \ref{pm4} and \ref{pm5} using MATLAB's \textit{vpasolve} routine. The output is the deformation state $\theta$ of the pouch motor and the bending state $q$ of the vine robot.
The relative bending $q/L_0$ is our primary metric for comparison to experimental data.

\subsection{cPAM}
For the cPAM, we use the same model as for the pouch motor proposed by \citet{Niiyama2015PouchDesign} but without the volume correction because the cPAM behaves like an ideal pouch motor with unconstrained sides. In contrast to the pouch motor, which actuates against the atmospheric pressure $p_\text{atm}$ on both sides, the cPAM acts against the atmospheric pressure on the top side and against the vine robot body pressure $p_\text{vine}$ on the bottom side. Therefore, we take the arithmetic mean between $p_\text{atm}$ and $p_\text{vine}$. Calculating with relative pressures simplifies this because $p_\text{atm}\,=\,0$:
\begin{gather}
    F_\text{cpam} = L_0 w \left(p_\text{cpam} - \frac{p_\text{atm} + p_\text{vine}}{2}\right) \frac{\cos{\theta}}{\theta} \\
    = L_0 w \left(p_\text{cpam} - \frac{p_\text{vine}}{2}\right) \frac{\cos{\theta}}{\theta}
\end{gather}
The contracted length $L_1$ is again related to the initial length $L_0$ and the deformation state $\theta$ of the cPAM.
\begin{gather}
    L_1 = L_0 \frac{\sin(\theta)}{\theta}
\end{gather}
Due to the integration of the cPAM into the vine robot body, the bending axis is lowered by the distance $s$. It depends on $D_\text{vine}$ and the width $w$ of the cPAM:
\begin{gather}
    D_\text{vine} = \frac{4 s^2 + w^2}{4s}
\end{gather}
Similar to the pouch motor, we have one equation for the force equilibrium and one equation for the geometric state. We input the cPAM pressure $p_\text{cpam}$ and solve for $q$:
\begin{gather}
    F_\text{vine} \sin(q) = F_\text{cpam} \sin\left(\frac{q}{2}\right) \\
    L_0 = \frac{q L_1} {2 \sin\left(\frac{q}{2}\right)} + q (D_\text{vine}-s)
\end{gather}

\subsection{fPAM}
The fPAM model is based on the linear contraction model introduced by \citet{Naclerio2020SimpleMuscle}. It requires the fPAM radius $r$ and the parameters $\alpha_0$, $a$, and $b$. We get the equations and the maximum contraction $\epsilon_\text{max} = 0.308$ from \citet{Naclerio2020SimpleMuscle}:
\begin{gather}
    r = \frac{w}{\pi}\\
    \alpha_0 = -a \sin \frac{\sqrt{\epsilon_\text{max}^2 - 2 \epsilon_\text{max} + 2/3}}{\epsilon_\text{max} - 1}\\
    a = 3 / \tan^2 \alpha_0\\
    b = 1 / \sin^2 \alpha_0
\end{gather}
The model's complexity arises from the pre-stretch of the fPAM when attached to the vine robot body. Therefore, the actuator force $F_\text{fpam}$ depends on both a pneumatic $F_\text{pneumatic}$ and elastic component $F_\text{elastic}$ (Figure \ref{fig:modeling}.F). 
When attaching the fPAM to a vine robot, we can only pre-stretch it to a certain amount $\epsilon_\text{ps}$. The relevant contraction for bending $\epsilon_\text{bend}$ can then be related to the initial $L_0$ and contracted $L_1$ length:
\begin{gather}
    \epsilon_\text{bend} = \epsilon - \epsilon_\text{ps} = \frac{L_0 - L_1}{L_0}
\end{gather}
We use the general contraction $\epsilon$ for calculation.
At $\epsilon_0$ the fPAM has no more pre-stretch, resulting in the definition of the elastic contraction $\epsilon_\text{elastic}$ relevant for $F_\text{elastic}$:
\begin{gather}
    \epsilon_\text{elastic} = \epsilon_\text{0} - \epsilon
\end{gather}
We performed a manual model parameter fitting to find the parameters $\epsilon_\text{ps}$ and $\epsilon_{0}$. This resulted in $\epsilon_{0} = 0.275$ for all fPAM actuators. The resulting values for $\epsilon_\text{ps}$ were 0.203, 0.198, and 0.207 for fPAM widths of 20, 40, and 60\,mm, respectively.
In addition, we determined the product $Et$ of the elasticity module $E$ and the thickness of the material $t$. At $\epsilon_\text{ps}$, the fPAM is fully pre-stretched and not inflated, so the fPAM force consists only of its elastic component. By setting this force equal to the vine robot force $F_\text{vine}$, we can solve for $Et$:
\begin{gather}
    F_\text{vine} = F_\text{fpam}(\epsilon_\text{ps}) = F_\text{elastic}(\epsilon_\text{ps}) = 2 \pi r E t (\epsilon_0-\epsilon_\text{ps})
\end{gather}
The resulting values for $Et$ were 3046\,N/m, 1421\,N/m, and 1084\,N/m for fPAM widths of 20, 40, and 60\,mm, respectively. 
Using these values, we can express the fPAM force $F_\text{fpam}$ in terms of the pneumatic $F_\text{pneumatic}$ and the elastic $F_\text{elastic}$ component, as introduced by \citet{Naclerio2020SimpleMuscle}:
\begin{gather}
    F_\text{pneumatic} = \pi r^2 [a(1-\epsilon)^2 - b] p_\text{fpam}\\
    F_\text{elastic} = 2 \pi r E t \epsilon_\text{elastic} = 2 \pi r E t (\epsilon_0 - \epsilon)\\
    F_\text{fpam} = \begin{cases}
            F_\text{pneumatic} + F_\text{elastic} & \epsilon_\text{ps} < \epsilon < \epsilon_0\\
            F_\text{pneumatic} & \epsilon_0 < \epsilon < \epsilon_\text{max}
        \end{cases}
\end{gather}
Because the fPAM continuously follows the shape of the vine robot, we equate the forces:
\begin{gather}
    F_\text{vine} = F_\text{fpam}
\end{gather}
Based on the input pressure $p_\text{fpam}$, we calculate the resulting contraction $\epsilon$ of the fPAM. Finally, we can relate $\epsilon$ to the relative bending $q/L_0$ of the vine robot:
\begin{gather}
    q/L_0 = \frac{\epsilon_\text{bend}}{\frac{1}{2}D_\text{vine}+ r} =  \frac{(\epsilon-\epsilon_\text{ps})}{\frac{1}{2}D_\text{vine} + r}
\end{gather}

%% file: chapter_5_results.tex
\section{Results}
\label{Results}

\begin{table}[t]
\caption{Eversion pressure with deflated actuators ($p_\text{0}$) and with actuators inflated to 15\,kPa ($p_\text{15}$). `nf' indicates geometrically infeasible dimensions, `nb' indicates dimensions which do not bend.}
\label{table:eversion}
\small
\begin{center}
\setlength{\tabcolsep}{4pt}
\renewcommand{\arraystretch}{1.0}
\begin{tabular}{cc|cccccc}
\multicolumn{2}{c|}{Actuator Type}           & \multicolumn{6}{c}{Pressure (kPa)}                                                                         \\ \cline{3-8} 
\multicolumn{1}{l}{} & \multicolumn{1}{l|}{}  & $p_\text{0}$   & \multicolumn{1}{c|}{$p_\text{15}$}   & $p_\text{0}$   & \multicolumn{1}{c|}{$p_\text{15}$}    & $p_\text{0}$    & $p_\text{15}$    \\ \hline
\hline
 Pouch Motor        &   $w$ (mm)          & \multicolumn{2}{c|}{20}                 & \multicolumn{2}{c|}{40}                 & \multicolumn{2}{c}{60} \\ \cline{2-8} 
                     & 20                     & 1.30  & \multicolumn{1}{c|}{1.14}   & 1.32 & \multicolumn{1}{c|}{1.32}   & 1.42  & 1.84   \\
$L_0$ (mm)               & 40                    & \multicolumn{2}{c|}{nb}   &  1.26 & \multicolumn{1}{c|}{1.52}   & 1.54  & 1.92   \\
                     & 60                    & \multicolumn{2}{c|}{nb}    & 1.26 & \multicolumn{1}{c|}{1.70}   & 1.76  & 1.72   \\ \hline
\hline                    
cPAM                 &    $w$ (mm)         & \multicolumn{2}{c|}{20}                 & \multicolumn{2}{c|}{40}                 & \multicolumn{2}{c}{60} \\ \cline{2-8} 
                     & 20                     & 1.22 & \multicolumn{1}{c|}{1.62}  & 1.62 & \multicolumn{1}{c|}{2.16}   & 1.56  & 2.32   \\
$L_0$ (mm)               & 40                    & \multicolumn{2}{c|}{nf}    & 1.56 & \multicolumn{1}{c|}{1.56}   & 1.70  & 2.74  \\
                     & 60                    & \multicolumn{2}{c|}{nf}   & \multicolumn{2}{c|}{nf}    & 1.56  & 3.02   \\ \hline
\hline
fPAM                 &   $w$ (mm)           & \multicolumn{2}{c|}{20}                 & \multicolumn{2}{c|}{40}                 & \multicolumn{2}{c}{60} \\ \cline{2-8} 
                     &                        & 0.60 & \multicolumn{1}{c|}{0.66}  & 0.60 & \multicolumn{1}{c|}{1.26}  & 0.66  & 1.44  
\end{tabular}
\end{center}
\end{table}

\begin{table}[t]
\caption{Normalized model error $e$ as defined in Section \ref{sec:model_results} for all three actuator types and all dimensions. W indicates the width and L indicates the length of an actuator.
}
\label{table:model_error}
\small
\begin{center}
\setlength{\tabcolsep}{5pt}
\renewcommand{\arraystretch}{1.0}
\begin{tabular}{cc|cc|cc}
\multicolumn{2}{c|}{Pouch Motor} & \multicolumn{2}{c|}{cPAM} & \multicolumn{2}{c}{fPAM}                      \\ \hline
\hline
Size (mm)      & $e$          & Size (mm)   & $e$      & Size (mm)       & $e$                      \\ \hline
W20xL20           & 0.177      & W20x L20        & 0.483  & W20                  & 0.034                  \\
W40xL20           & 0.205      & W40x L20        & 0.255  & W40 & 0.019 \\
W40xL40           & 0.196     & W40x L40        & 0.169  &                      &                        \\
W40xL60           & 0.797      &                  &        &                      &                        \\
W60xL20           & 0.132      & W60x L20        & 0.435  & W60 & 0.170  \\
W60xL40           & 0.199      & W60x L40        & 0.113  &                      &                        \\
W60xL60           & 0.355      & W60x L60        & 0.102  &                      &     
\end{tabular}
\end{center}
\end{table}

\subsection{Eversion}
Table \ref{table:eversion} presents the pressure measurements required for eversion when all pouches are inflated to 0 and 15\,kPa. The fPAM has a much lower eversion pressure. This is because the fPAM uses a thin and flexible, low-friction, 30D silicone-coated ripstop nylon, whereas the pouch motor and cPAM use a stiffer 70D TPU-coated ripstop nylon. When the actuators are not inflated, the pouch motor and cPAM require a similar eversion pressure, suggesting that the eversion pressure mainly depends on the material properties. The maximum eversion pressure required for any deflated actuator was 1.76\,kPa. When the actuators are inflated, actuators with larger dimensions require a higher eversion pressure. As a result, a vine robot with cPAMs requires the highest eversion pressure. The maximum eversion pressure required for any inflated actuator was 3.02\,kPa.

\begin{figure}[t]
    \centering
    \includegraphics[width=0.5\textwidth]{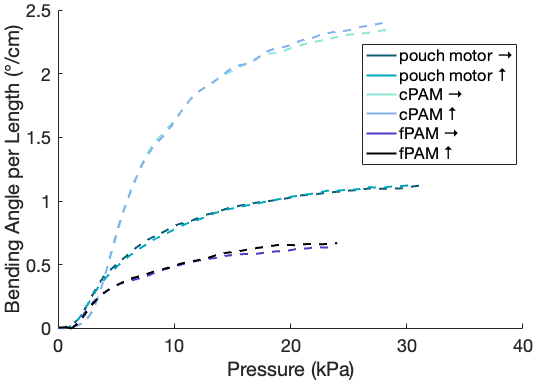}
    \caption{Comparison of the quasi-static bending performance in horizontal (→, without gravity) and vertical (↑, with gravity) orientation. Dashed lines indicate the mean over five iterations for the largest dimensions of each actuator type: W60xL60\,mm pouch motor, W60xL60\,mm cPAM, and W60\,mm fPAM.}
    \label{fig:results_orientation}
\end{figure}

\begin{figure*}[t]
    \centering
   \includegraphics[width=\linewidth]{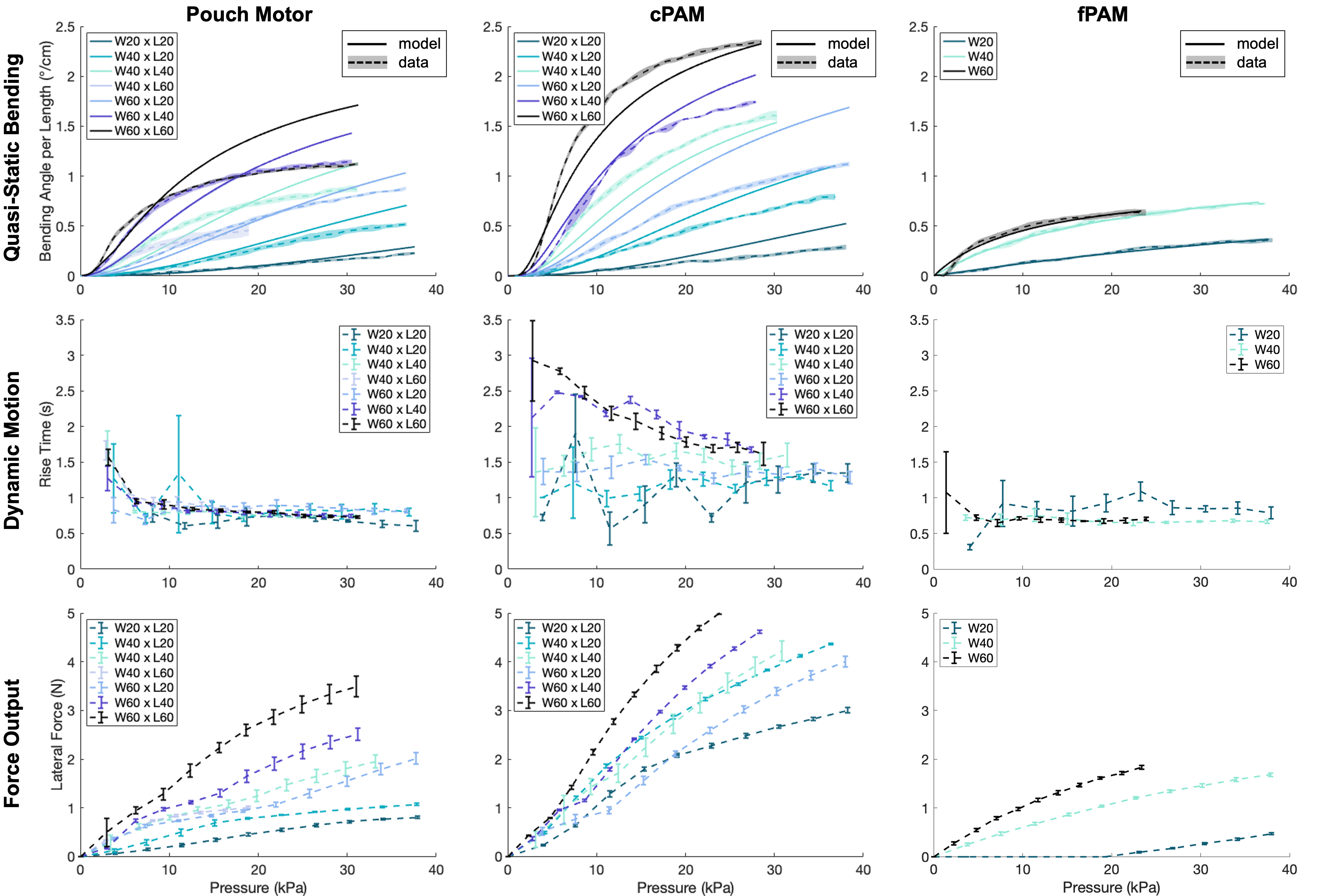}
    \caption{Experimental results for the three actuator types with respect to the inflation pressure (x-axis). W indicates the width and L indicates the length of an actuator. Left to right: Pouch motor, cPAM, and fPAM results. Top to bottom: Quasi-static bending results characterized by the bending angle per nominal length, dynamic motion capabilities characterized by the 10-90\% rise time, and the resulting lateral force when the tip of the vine robot is constrained. In the first row, the mean of five experimental measurements is shown by dashed lines, the experimental standard deviation is shown by the shaded areas, and the analytical model is shown by solid lines. In the second and third rows, dashed lines show the mean over five iterations and the error bars indicate the corresponding standard deviation.}
    \label{fig:results}
\end{figure*}

\begin{figure*}[t]
    \centering
    \includegraphics[width=\textwidth]{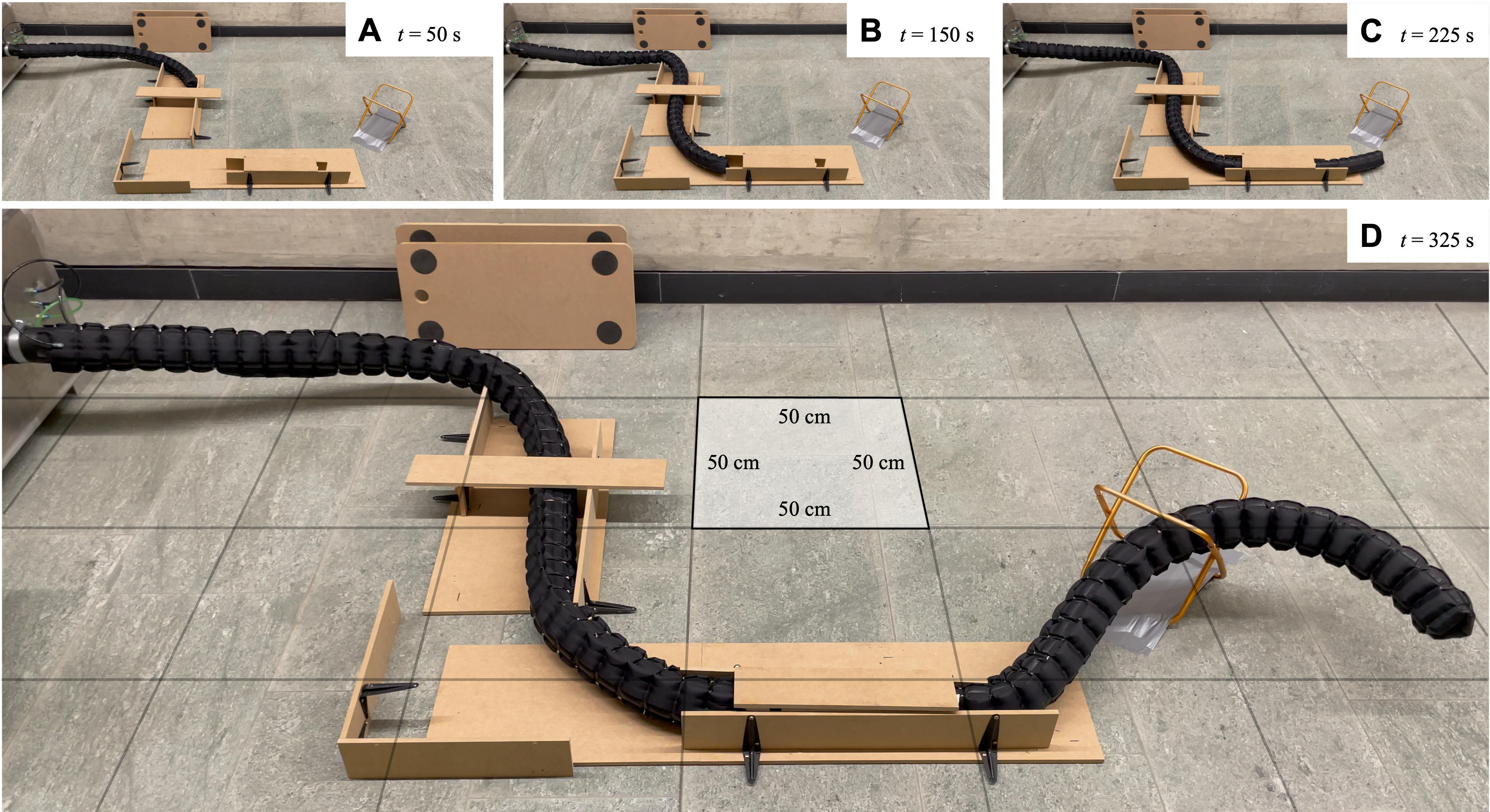}
    \caption{Demonstration of navigating an obstacle course with a 4.8\,m long vine robot of 80\,mm diameter and three strands of W60xL60\,mm cPAMs for steering in the three-dimensional space. The vine robot is supported by an improved version of the supply box shown in \citet{DerMaur2021RoBoa:Applications}. A.-C. The vine robot with corresponding time stamps while moving through the obstacle course. D. The final state of the vine robot along a right turn, passage under a bridge, left turn, passage with a shrunken diameter, and a vertical turn. The grid indicates dimensions of 50x50\,cm. The dimensions of the demonstrated vine robot are derived from the results of the quasi-static bending experiments.}
    \label{fig:demo}
\end{figure*}

\subsection{Quasi-Static Bending}
\label{bending_results}
Figure \ref{fig:results_orientation} shows the quasi-static bending for the largest dimensions of the three actuator types with and without gravity. The results are very similar, indicating that gravity does not influence the bending behavior due to the lightweight structure of the vine robot. This result holds true for other dimensions and dynamic bending. However, for longer vine robots or when a tip mount is attached to the front, bending against gravity will become more difficult. The force capabilities of the vine robot will then become crucial.
Figure \ref{fig:results} (first row) displays the experimental pressure-to-bending relationship and the analytical model prediction without gravity (horizontal configuration). The best bending performance was achieved by the cPAM, which almost reaches the theoretical maximum value ($2.6^\circ/\text{cm}$ for a vine robot with 80\,mm diameter \cite{Kubler2022ASteering}). The integration of the cPAM into the vine robot body and the folded material that does not constrain the sides are critical for achieving this performance. 
The pouch motor bends slightly more than the fPAM.
Larger dimensions result in higher bending for both the pouch motor and cPAM, with both the width and the length being essential. The cPAM significantly outperforms the pouch motor for large dimensions, but the performance at small dimensions is similar. This suggests that the fold structure requires a certain size to work.
The fPAM shows a similar performance for $w=40\,\text{mm}$ and $w=60\,\text{mm}$, better than for $w=20\,\text{mm}$. 
This suggests that the fPAM requires a certain dimension to achieve higher bends, or that small fPAMs require a much higher input pressure. Furthermore, the stiffness of the glued connection and the pre-stretch have a large impact on the bending performance.

\subsection{Analytical Pressure-to-Bending Model}
\label{sec:model_results}
To analyze the pressure-to-bending models, we define the average normalized error $e$ between the bending predicted by the analytical model $q_\text{model}(p_i)$ and the experimental bending data $q_\text{data}(p_i)$. The first data points with very low pressure and bending are neglected to avoid a normalization by values close to zero:
\begin{gather}
    e = \frac{1}{n}\sum_{i=1}^{n} \frac{|q_\text{model}(p_i) - q_\text{data}(p_i)|}{q_\text{data}(p_i)}
\end{gather}
Figure \ref{fig:results} shows the predicted bending in the first row and Table \ref{table:model_error} reports the error $e$.
The analytical model can distinguish between different actuator dimensions. It performs best for small pouch motors, large cPAMs, and the fPAM. However, the model fails to predict the convergence for large pouch motors because the ideal model always converges to the theoretical maximum. The volume multiplier does not change the convergence value but only slows the increase with respect to the actuator pressure. The cPAM model overpredicts bending for small dimensions. This is likely because the cPAMs do not fully unfold at small dimensions, although the model assumes exactly this. The geometric model does not account for material stiffness, which can hinder small cPAMs from unfolding. The fPAM model works well for all dimensions. It requires the calculation of actuator-specific parameters for each dimension, resulting in a tuning of the model. The pouch motor and cPAM models work without any parameter identification or tuning.

\subsection{Dynamic Motion}
The second row in Figure \ref{fig:results} depicts the dynamic bending behavior, measured by the rise time resulting from a step input to a specific pressure. 
The fPAM actuates fastest, closely followed by the pouch motor. The fPAM can actuate quickly because it is not segmented and therefore does not experience a constriction in airflow. The pouch motor has segments but a smaller volume to inflate.
The cPAM performs poorly because it has segments and a large volume due to its folded structure. 
The test may not fully showcase the capability of the fPAM. \citet{Naclerio2020SimpleMuscle} showed that the fPAM can outperform other pneumatic actuators in a frequency response test. The dynamic motion depends not only on the actuator but also on the air supply system. We used QB3 pressure regulators that are restricted in their airflow.

\subsection{Force Output}
The last row in Figure \ref{fig:results} presents the results of the lateral force output. The cPAM demonstrates the highest force output as the force is related to the area of the actuator. Additionally, the welded integration of cPAM with the vine robot results in a sturdy and robust connection. The pouch motor can produce a force smaller than the cPAM but larger than the fPAM. Larger dimensions within an actuator type generate higher force outputs at the same pressure due to their larger area.

%% file: chapter_6_demonstrator.tex
\section{Demonstration} \label{Demonstrator}
To showcase the possibilities of a vine robot with improved actuators, specifically an optimized cPAM, we manufactured a vine robot with a length of 4.8\,m to traverse a tortuous obstacle course (Figure \ref{fig:demo}), making the bending ability of the robot critical for its success. The vine robot has a diameter of 80\,mm and cPAMs with a width and length of 60\,mm, which, based on the results in Section \ref{bending_results}, produce the most significant bending. The vine robot has three lines of actuators, allowing for steering in the three-dimensional space. We used an improved version of the supply box introduced by \citet{DerMaur2021RoBoa:Applications}. Due to material availability, we changed the material from ripstop nylon to pure 70 Denier nylon with a one-sided TPU coating (Extremtextil e.K., Dresden, Germany). We manufactured the complete vine robot in two days using a high-frequency welding machine (Walser Kunststoffwerk AG, Bürglen, Switzerland).\\
The vine robot moved through the obstacle course with an average speed of 1.5\,cm/s as commanded by the user. The robot's teleoperated movement, using a PlayStation controller, slowed when it had to target a specific obstacle or opening. However, in free space without the need for precise steering, the vine robot can move faster. The robot successfully navigated around a $90\,^\circ$ right turn, passed under a bridge, and performed a $90\,^\circ$ left turn with a bend of approximately $2.0\,^\circ/\text{cm}$. The robot then moved through a tunnel, shrinking its body to 6\,cm in diameter, followed by lifting itself over 15\,cm to pass through the final opening.\\
This test demonstrates the high maneuverability of a compliant vine robot with improved actuators in a three-dimensional obstacle course.

%% file: chapter_7_conclusion.tex
\section{Conclusion} \label{Conclusion}

We compared three commonly used soft pneumatic actuators: the pouch motor, the cylindrical pneumatic artificial muscle (cPAM), and the fabric pneumatic artificial muscle (fPAM). These actuators find applications in soft continuum robots, such as soft growing vine robots.

We developed a testing methodology and setup to measure a set of performance parameters: eversion, quasi-static bending, dynamic motion, and force output. The actuators were attached to a vine robot, and different sizes were tested for each actuator. The pouch motor is advantageous for prototyping because its simple rectangular structure enables fast fabrication and attachment to the vine robot. The cPAM extends the pouch motor concept by adding folded material on the sides, making it behave like an ideal pouch motor that can inflate to form a complete cylinder. This improves its bending capability, making it the best actuator to perform strong bends and navigate tight turns. Due to its large volume and strong connection to the vine robot, it can generate the highest lateral force. The fPAM requires the lowest pressure to evert because of its thin and low-friction material. It can perform dynamic motions best as it consists of a single tube without separated air chambers.

When comparing different dimensions of the actuators, we found that larger actuators generate more significant bends and forces, whereas smaller actuators react faster and require a lower eversion pressure. Both length and width should be maximized for large pouch motors and cPAMs.

The analytical pressure-to-bending models assume a static force equilibrium between the vine robot and the pneumatic actuator, accounting for the unique geometric conditions of each actuator type. They distinguish between actuators of different dimensions and correctly predict the performance order. The models predict the bending particularly well for the fPAM, small pouch motors, and large cPAMs, but overpredict the bending of large pouch motors and small cPAMs.

The experimental methods and models in this work can be used to design, test, and optimize future actuators for vine robots and soft continuum robots. More efficient and repeatable fabrication approaches will improve the vine robot design and performance, and the analytical model can enhance future control approaches.